\documentclass[11pt,a4paper]{article}
\newcommand{\olidweak}{{\textit{SOLID}}}
\newcommand{\olid}{{\textit{OLID}}}
\usepackage[hyperref]{acl2021}
\usepackage{times}
\usepackage{latexsym}

\usepackage{amsfonts}
\usepackage{microtype}
\usepackage{multirow}
\usepackage{enumitem}
\usepackage{booktabs}       
\usepackage{hhline}
\usepackage{amsmath}
\usepackage{graphicx}
\usepackage{subcaption}

\usepackage{lipsum}
\newcommand\blfootnote[1]{%
  \begingroup
  \renewcommand\thefootnote{}\footnote{#1}%
  \addtocounter{footnote}{-1}%
  \endgroup
}
\usepackage{xurl}
\usepackage[T1]{fontenc}

\usepackage[utf8]{inputenc}

\usepackage{microtype}

\usepackage{microtype}

\aclfinalcopy 
\def\aclpaperid{655} 

\title{SOLID: A Large-Scale Semi-Supervised Dataset\\ for Offensive Language Identification}

\author{Sara Rosenthal\textsuperscript{1}, Pepa Atanasova\textsuperscript{2}, Georgi Karadzhov\textsuperscript{3}, \\ {\bf Marcos Zampieri\textsuperscript{4}, Preslav Nakov\textsuperscript{5}}\\
\textsuperscript{1}IBM Research, USA, 
\textsuperscript{2}University of Copenhagen, Denmark, \\
\textsuperscript{3}University of Cambridge, UK,
\textsuperscript{4}Rochester Institute of Technology, USA, \\
\textsuperscript{5}Qatar Computing Research Institute, HBKU, Qatar\\
{\tt sjrosenthal@us.ibm.com} }

\begin{document}
\maketitle
\begin{abstract}

The widespread use of offensive content in social media has led to an abundance of research in detecting language such as hate speech, cyberbullying, and cyber-aggression. Recent work presented the \olid{} dataset, which follows a taxonomy for offensive language identification that provides meaningful information for understanding the type and the target of offensive messages. However, it is limited in size and it might be biased towards offensive language as it was collected using keywords. In this work, we present \olidweak{}, an expanded dataset, where the tweets were collected in a more principled manner. \olidweak{} contains over nine million English tweets labeled in a semi-supervised fashion. We demonstrate that using \olidweak{} along with \olid{} yields sizable performance gains on the \olid{} test set for two different models, especially for the lower levels of the taxonomy. 
\end{abstract}

\section{Introduction}
\blfootnote{WARNING: This paper contains tweet examples and words that are offensive in nature.}
Offensive language in social media has become a concern for governments, online communities, and social media platforms. Free speech is an important right, but moderation is needed in order to avoid unexpected serious repercussions. In fact, this is so serious that many countries have passed or are planning legislation that makes platforms responsible for their content, e.g.,~the \textit{Online Harm Bill} in the UK and the \textit{Digital Services Act} in the EU. Even in USA, content moderation, or the lack thereof, can have significant impact on businesses (e.g.,~Parler was denied server space), on governments (e.g.,~the U.S. Capitol Riots), and on individuals (e.g.,~hate speech is linked to self-harm). As human moderators cannot cope with the volume, there is a need for automatic systems that can assist them. 

There have been several areas of research in the detection of offensive language \cite{basile-etal-2019-semeval,fortuna2018survey,ranasinghe-zampieri-2020-multilingual}, covering overlapping characteristics such as toxicity, hate speech, cyberbullying, and cyber-aggression. Moreover, it was proposed to use a hierarchical approach to analyze different aspects, such as the type and the target of the offense, which helps provide explainability. The Offensive Language Identification Dataset, or \olid, \cite{zampieri-etal-2019-predicting} is one such example, and it has been widely used in research. \olid{} contains 14,100 English tweets, which were manually annotated using a three-level taxonomy:

\textbf{A}: Offensive Language Detection 

\textbf{B}: Categorization of Offensive Language

\textbf{C}: Offensive Language Target Identification

\noindent This taxonomy makes it possible to represent different kinds of offensive content as a function of the {\em type} and the {\em target}. For example, offensive messages targeting a group are likely to be hate speech, whereas such targeting an individual are probably cyberbullying.
The taxonomy was also used for languages such as Arabic \cite{mubarak-etal-2021-arabic}, and Greek \cite{pitenis-etal-2020-offensive}, allowing for multilingual learning and analysis. 

An inherent feature of the hierarchical annotation is that the lower levels of the taxonomy contain a subset of the instances in the higher levels, and thus there are fewer instances in the categories in each subsequent level. This makes it very difficult to train robust deep learning models on such datasets. Moreover, due to the natural infrequency of offensive language (e.g.,~less than 3\% of the tweets are offensive when selected at random), obtaining offensive content is a costly and time-consuming effort. Here, we address these limitations by proposing a new dataset: \textbf{S}emi-Supervised \textbf{O}ffensive \textbf{L}anguage \textbf{I}dentification \textbf{D}ataset (\olidweak{}). 

\noindent Our contributions are as follows:
\begin{enumerate}
    \item We are the first to apply a semi-supervised method for collecting new offensive data using \olid{} as a seed dataset, thus avoiding the need for time-consuming annotation. 
    \item We create and publicly release \olidweak{}, a training dataset containing 9 million English tweets for offensive language identification, the largest dataset for this task.\footnote{Available at: \url{http://sites.google.com/site/offensevalsharedtask/solid}} \olidweak{} is the official dataset of the SemEval shared task OffensEval-2020 \cite{zampieri-etal-2020-semeval}.
    \item We demonstrate sizeable improvements over prior work on the middle and the lower levels of the taxonomy, where gold training data is scarce, when training on \olidweak{} and testing on \olid{}.
    \item We provide a new larger test set and a comprehensive analysis of \textit{EASY} (i.e.,~simple explicit tweets such as using curse words) and \textit{HARD} (i.e.,~more implicit tweets that use underhanded comments or racial slurs) examples of offensive tweets.
\end{enumerate}

\noindent The remainder of this paper is organized as follows: Section~\ref{sec:RW} presents related studies in aggression identification, cyberbullying detection, and other related tasks. Section~\ref{sec:olid} describes the \olid{} dataset and the annotation taxonomy. Section~\ref{S:models} introduces our computational models. Section~\ref{S:dataset} presents the \olidweak{} dataset. Section~\ref{sec:results} discusses the experimental results and Section~\ref{sec:discussion} offers additional discussion and analysis. Finally, Section~\ref{sec:conclusion} concludes and discusses possible directions for future work.

\section{Related Work}
\label{sec:RW}

There have been several recent studies on offensive language, hate speech, cyberbulling, aggression, and toxic comment detection. See \cite{Survey:2021:Abusive:Language} for an overview.

Hate speech identification is by far the most studied abusive language detection task \cite{ousidhoum-etal-2019-multilingual,chung-etal-2019-conan,mathew2020hatexplain}. One of the most widely used datasets is the one by \citet{davidson2017automated}, which contains over 24,000 English tweets labeled as non-offensive, hate speech, and profanity. A recent shared task on this topic is HatEval \cite{basile-etal-2019-semeval}. 

For cyberbullying detection, \citet{xu-etal-2012-learning} used sentiment analysis and topic models to identify relevant topics. \citet{dadvar2013improving} and \citet{safi-samghabadi-etal-2020-detecting} studied utility of the conversational context. In particular, \citet{dadvar2013improving} used user-related features such as the frequency of profanity in the previous messages. 
More recent work has addressed the issues of scalable and timely detection of cyberbullying in online social networks. To this end, \citet{10.1145/3167132.3167317} used a dynamic priority scheduler, and \citet{10.1145/3308558.3313462} proposed a sequential hypothesis testing. \citet{safi-samghabadi-etal-2020-detecting} constructed a dataset of cyberbullying episodes from the semi-anonymous social network ask.fm.

There were two editions of the TRAC shared task on Aggression Identification \cite{kumar-etal-2018-benchmarking,kumar-etal-2020-evaluating}, which provided participants with datasets containing annotated Facebook posts and comments in English and Hindi for training and validation. Then, Facebook and Twitter datasets were used for testing. The goal was to discriminate between three classes: non-aggressive, covertly aggressive, and overly aggressive. Two other shared tasks addressed toxic language. The Toxic Comment Classification Challenge\footnote{\url{http://kaggle.com/c/jigsaw-toxic-comment-classification-challenge}} at Kaggle provided participants with comments from Wikipedia annotated using six labels: toxic, severe toxic, obscene, threat, insult, and identity hate. The recent SemEval-2021 Toxic Spans Detection shared task addressed the identification of the token spans that made a post toxic \cite{pav2021semeval}.

There were several shared tasks that have focused specifically on offensive language identification, e.g.,~GermEval-2018~\cite{wiegand2018overview}, which focused on offensive language identification in German tweets, HASOC-2019~\cite{mandl2019overview}, and TRAC-2018\cite{fortuna-etal-2018-merging}. 

In this paper, we extend the prior work of the \olid{} dataset~\cite{zampieri-etal-2019-predicting}. It is annotated using a hierarchical annotation schema as in~\cite{basile-etal-2019-semeval,mandl2019overview}. In contrast to prior approaches, it takes both the target and the type of offensive content into account. This allows multiple types of offensive content (e.g.,~hate speech and cyberbullying) to be represented in \olid's taxonomy. Herem we create a large-scale semi-supervised dataset using the same annotation taxonomy as in \olid.

\begin{table}[t]
\centering
\fontsize{8.5}{10}\selectfont{
\setlength{\tabcolsep}{1pt}
\begin{tabular}{p{54mm}lccc@{}}
\toprule
 \textbf{Tweet} & \textbf{A} & \textbf{B} & \textbf{C} \\ \midrule
 @USER Anyone care what that dirtbag says?& OFF & TIN & IND\\
 Poor sad liberals. No hope for them. & OFF & TIN & GRP\\
 LMAO....YOU SUCK NFL & OFF & TIN & OTH  \\
@USER What insanely ridiculous bullshit. & OFF & UNT & --\\
@USER you are also the king of taste & NOT & -- & -- \\ \bottomrule
\end{tabular}
\caption{Examples from the \olid{} dataset.}
\label{tab:olid_examples}
}
\end{table}

\section{The OLID Dataset}
\label{sec:olid}

The \olid{} \cite{zampieri-etal-2019-predicting} dataset tackles the challenge of detecting offensive language using a labeling schema that classifies each example using the following three-level hierarchy: 

\paragraph{Level A: \emph{Offensive Language Detection}}
Is the text offensive?

\noindent\textbf{OFF} Inappropriate language, insults, or threats.

\noindent\textbf{NOT} Neither offensive, nor profane.

\paragraph{Level B: \emph{Categorization of Offensive Language}} Is the offensive text targeted?

\noindent\textbf{TIN} Targeted insult or threat towards a group or an individual.

\noindent\textbf{UNT} Untargeted profanity or swearing.

\paragraph{Level C: \emph{Offensive Language Target Identification}} What is the target of the offense?

\noindent\textbf{IND} The target is an individual explicitly or implicitly mentioned in the conversation;

\noindent\textbf{GRP} Hate speech targeting a group of people based on ethnicity, gender, sexual orientation, religion, or other common characteristic.

\noindent\textbf{OTH} Targets that does not fall into the previous categories, e.g.,~organizations, events, and issues.

\vspace{3mm}

The taxonomy was successfully adopted for several languages \cite{mubarak-etal-2021-arabic,pitenis-etal-2020-offensive,sigurbergsson-derczynski-2020-offensive,coltekin-2020-corpus}, and it was used in a series of shared tasks \cite{zampieri-etal-2019-semeval,mandl2019overview}. Tweets from the \olid{} dataset labeled with the taxonomy are shown in Table~\ref{tab:olid_examples}. The \olid{} dataset consists of 13,241 training and 860 test tweets. 

Table~\ref{tab:training-data-distribution} presents detailed statistics about the distribution of the labels. There is a substantial class imbalance at each level of the annotation, especially at Level B. Moreover, there is a sizable difference in the total number of annotations between the levels due to the schema, e.g.,~Level C is 30\% smaller than Level A, and the data sizes for B and C are rather small. These drawbacks indicate the need to create a larger dataset.

\section{Models}
\label{S:models}

In this section, we describe the models used for semi-supervised annotation and for evaluating the contribution of \olidweak{} for offensive language identification.
We use a suite of heterogeneous machine learning models: PMI \cite{turney2003measuring}, FastText \cite{joulin-etal-2017-bag}, LSTM \cite{hochreiter1997long}, and BERT \cite{devlin-etal-2019-bert}, which have diverse inductive biases. This is an essential prerequisite for our semi-supervised setup (see Section~\ref{S:DemocraticCoTraining}), as we assume that an ensemble of models with different inductive biases would ecrease each individual model's bias.

\subsection{PMI}

We use a PMI-based model that computes the $n$-gram-based similarity of a tweet to the tweets of a particular class \textit{c} in the training dataset. The model is considered na\"{i}ve as it accounts only for the $n$-gram frequencies in the discrete token space and only in the context of $n$ neighboring tokens. We compute the PMI score~\cite{turney2003measuring} of each $n$-gram in the training set w.r.t. each class:
\begin{equation}
\small
PMI(w_i, c_j) = log_2\left(\frac{p(w_i, c_j)}{p(w_i)*p(c_j)}\right)
\end{equation}
\noindent where $p(w_i, c_j)$ is the frequency of $n$-gram $w_i$ in instances of class $c_j$, $p(w_i)$ is the frequency of $n$-gram $w_i$ in instances from the entire training dataset, and $p(c_j)$ is the frequency of class $c_j$. Additionally, we find that semantically oriented PMI scores~\cite{turney2003measuring} improve the performance of this na\"{i}ve method: 
\begin{equation}
\small
PMI-SO(w_i, c_j) = log_2(\frac{p(w_i, c_j) * p(C \setminus\{c_j\})}{p(w_i, C\setminus\{c_j\}) * p(c_j)})
\end{equation}
\noindent where $C\! \setminus\! \{c_j\}$ is the set of all classes except $c_j$. 

At training time, we collect the frequencies of the $n$-grams on the training set. At inference time, we use the frequencies to calculate PMI and PMI-SO scores for each unigram and bigram in each instance, and then we average PMI and PMI-SO into a single score for each instance and class. Finally, we select the class with the highest score. If the instance contains no words with associated scores, we choose NOT for Level A, UNT for Level B (i.e.,~the classes most likely to contain neutral orientation), and the majority class IND for Level C. We remove words appearing less than five times in the training set, and we add a smoothing of 0.01 to each frequency. 

\subsection{FastText}

A suitable extension to the word-based model is to use subword representations to overcome the naturally noisy structure of the tweets. FastText \cite{joulin-etal-2017-bag} is a subword model, which has shown strong performance on various tasks without the need for extensive hyper-parameter tuning. It uses a shallow neural model for text classification similar to the continuous bag-of-words model~\cite{mikolov2013}. However, instead of predicting the word based on its neighbors, it predicts the target label based on the sample's words. FastText offers a valuable, diverse modeling representation to the ensemble due to its differences with the simple PMI model and the heavy-lifting LSTM and BERT models.
We train FastText with bigrams and a learning rate of 0.01 for Levels A and B, and with trigrams and a learning rate of 0.09 for Level C. All tasks use a window size of five and a hierarchical softmax loss.

\subsection{LSTM}

Unlike the above models (PMI and FastText), an LSTM model~\cite{hochreiter1997long,vaswani2017attention} can account for long-distance relations between words. Our LSTM model has an embedding layer, which we initialize with a concatenation of 300-dimensional GloVe embeddings \cite{pennington-etal-2014-glove} and 300-dimensional FastText Common Crawl embeddings \cite{grave-etal-2018-learning}. Then, follow a dropout layer, followed by a bi-directional LSTM layer with an attention mechanism on top of it. Next, we concatenated the attention mechanism's output with averaged and maximum global poolings on the outputs of the LSTM model. The final prediction was produced by a sigmoid layer for Levels A and B, where we have a binary classification, and a by softmax layer for Level C, where we have three classes. 
We trained the LSTM model using early stopping with patience for no improvements over the validation loss of up to five epochs. 

In terms of dimensionality, for Level A, we used a hidden size of 128, a dropout rate of 0.3, a batch size of 256, and a learning rate of 0.0002. For levels B and C, we used a hidden size of 50, a dropout rate of 0.1, a batch size of 32, and a learning rate of 0.0001. Finally, we used the Adam optimizer for training.

\subsection{BERT}

Recently, the Transformer architecture~\cite{vaswani2017attention} has demonstrated state-of-the-art performance for several NLP tasks, offering both high representational power and robustness. Here, we exploit the benefits of transfer learning in a low-resource setting by using the pre-trained BERT model \cite{devlin-etal-2019-bert}, which we fine-tune for our tasks (i.e.,~classification for each of the three levels of the taxonomy). In our experiments, we use the base uncased BERT model implementation from HuggingFace, which has 12 layers, a hidden size of 768, and 12 attention heads, amounting to 110 million parameters. We then fine-tune the model for 2, 3, and 3 epochs for Levels A, B, and C, respectively. We use learning rates of 0.00002 for Levels A and B, and 0.00004 for Level C. We apply per-class weights to cope with the data imbalance in Level C as follows: IND=1, GRP=2, OTH=10. We use the Adam optimizer and a linear warm-up schedule with a 0.05 warm-up ratio.

\subsection{Democratic Co-training}
\label{S:DemocraticCoTraining}

Democratic co-training \cite{zhou2004democratic} is a semi-supervised technique, commonly used to create large datasets with noisy labels when provided with a set of diverse models trained in a supervised way. It has been successfully applied in tasks like time series prediction with missing data \cite{mohamed2007co}, early prognosis of academic performance \cite{kostopoulos2019multiview}, as well as for tasks in the health domain \cite{longstaff2010improving}. In our case, we use models with diverse inductive biases to label the target tweet, which can help ameliorate the individual model biases, thus yielding predictions with a lower degree of noise.

In particular, we use democratic co-training to generate semi-supervised labels for all three levels of the \olidweak{} dataset, using \olid{} as a seed dataset, and applying distant supervision using an ensemble of the above-described models as follows:

\begin{enumerate}
\item Train $N$ \textit{diverse} supervised models $\{M_j(X)\}$, where $j \in [1, N]$ on a dataset with gold labels $X=\{(x_i, y_i)\}$, where $i \in [1, |X|]$ 
\item For each example $x'_i$ in the unannotated dataset $X' = \{(x_i')\}$, $|i \in [1, |X'|])$ and each model $M_j$, predict the confidence ${p'_i}^j$ for the positive class.
\end{enumerate}

\begin{table}[t]
\centering
\small
\setlength{\tabcolsep}{4.2pt} 
\begin{tabular}{ccrrrr}
\toprule
\multirow{2}{*}{\textbf{Level}} &\multirow{2}{*}{\textbf{Label}} & \multicolumn{2}{c}{\bf \olid} & \multicolumn{2}{c}{\bf \olidweak} \\
& & \bf Train & \bf Test & \bf Train & \bf Test \\ 
\midrule
\multirow{2}{*}{\textbf{A}} & OFF & $4{,}640$ & $240$ &  $1{,}448{,}861$ & $3{,}002$ \\
& NOT & $9{,}460$ & $620$ & $7{,}640{,}279$ & $2{,}991$ \\
\midrule
\multirow{2}{*}{\textbf{B}} & TIN & $4{,}089$ & $213$ & $149{,}550$ & $1{,}546$ \\
& UNT & 551 & $27$ & $39{,}424$ & $1{,}451$ \\
\midrule
\multirow{3}{*}{\textbf{C}} & IND & $2{,}507$ & $100$ & $120{,}330$ & $1{,}055$ \\
& GRP & $1{,}152$ & $78$ & $22{,}176$ & $349$ \\
& OTH & $430$ & $35$ & $7{,}043$ & $140$ \\
\bottomrule 
\end{tabular}
\caption{Statistics about the training and the testing data distribution for the \olid{} and the \olidweak{} datasets.}
\label{tab:training-data-distribution}
\end{table}

\section{The \olidweak{} Dataset}
\label{S:dataset}

In this section, we describe the process of collecting and annotating data for \olidweak{}. We collect a large set of over 12 million tweets, and we labeled nine million of them using the democratic co-training approach described in the previous section. Table~\ref{tab:training-data-distribution} shows some statistics about the resulting dataset for each level of the taxonomy.

\subsection{A Large-Scale Dataset of Tweets}
\label{sec:data:tweets}

We collected our data in 2019 from Twitter using the Twitter streaming API\footnote{\url{http://developer.twitter.com/en/docs}} and Twython\footnote{\url{http://twython.readthedocs.io}}.
We searched the API using the twenty most common English stopwords (e.g.,~\emph{the}, \emph{of}, \emph{and}, \emph{to}) to ensure truly random tweets and to avoid rate limits imposed by the Twitter platform. Using stopwords ensured that we were more likely to obtain English tweets as well as a diverse set of random tweets. We kept the stream tweet collection running the entire time and we continuously chose a stopword at random based on its frequency in Project Gutenberg, a sizeable monolingual corpus. For each query, we collected 1,000 tweets per stopword. Thus, frequent stopwords were used more frequently to collect tweets. A full list of the stopwords and their frequency is shown in Appendix~\ref{App:A}.
We used this data collection approach in an attempt to try to help mitigate the biases found in \olid{}. \olid{} was collected using a predefined list of keywords that were more likely to retrieve offensive tweets, which caused offensive tweets in \olid{} to be explicit and easier to classify. 
In contrast, in our case here, the tweets we collected for \olidweak{} contain implicit and explicit offensive text. This allows us to study the performance of various models in hard classification cases.

\begin{table}[t]
\centering
\small
\begin{tabular}{lccc}
\toprule
\textbf{Model} & \textbf{Level A} & \textbf{Level B} & \textbf{Level C} \\\midrule
Majority Baseline & 0.419 & 0.470 & 0.214 \\
\midrule
BERT & 0.816 & 0.705& 0.568 \\
PMI & 0.684 & 0.498 & 0.461 \\
LSTM & 0.681 & 0.657 & 0.585\\
FastText & 0.662 & 0.470 & 0.590\\
\bottomrule
\end{tabular}
\caption{Macro-F1 score of the models in the democratic co-training ensemble on the \olid{} test set.}
\label{tab:ensemble}
\end{table}

We used the \texttt{langdetect} tool\footnote{\url{http://pypi.org/project/langdetect/}} to select English tweets, and we discarded tweets with less than 18 characters or such that were less than two words long. We substituted all user mentions with \texttt{@USER} for anonymization purposes. We also ignored tweets with URLs as such that did not tend to be offensive and might be less self-contained, e.g.,~that could have a link to an article, an image, a video, etc. Understanding such tweets would require going beyond their purely textual content. In total, we collected over twelve million tweets. We kept nine million as training data, and we created a new test set from a portion of the remaining three million tweets.

\subsection{Semi-Supervised Training Dataset}
\label{sec:semisuptraining}

We used the democratic co-training setup described in Subsection~\ref{S:DemocraticCoTraining} to create the semi-supervised dataset. We first trained each model on the \olid{} dataset using 10\% of the training dataset for validation. The performance of the individual models on the \olid{} dataset is shown in Table~\ref{tab:ensemble}. We can see that BERT is the best model for Level A, and that PMI performs almost on par with the LSTM model. We believe that this is due to the size of the dataset and to the fact that a simple lexicon of curse words would be highly predictive of the offensive content present in a tweet. The performance of the FastText model is the lowest by 2 points. 

For Level B, BERT performs best, followed by the LSTM model. The task is more challenging at this level for frequency and $n$-gram-based approaches such as PMI and FastText.

Finally, the overall performance of the models at Level C decreases further. This is expected as the size of the dataset becomes smaller, and the task becomes one of three-way classification, whereas Levels A and B are two-way. Here, BERT and LSTM outperform FastText and PMI, with BERT being the best model.

\begin{table*}[t]
\centering
\setlength{\tabcolsep}{3pt} 
\small
\begin{tabular}{cp{72mm}cccccccc}
\toprule
\bf Level & \bf Examples & \bf BERT & \bf LSTM & \bf FT & \bf PMI & \bf AVG & \bf STD & \bf Label & \bf E/H \\ \midrule
\multirow{3}{*}{\textbf{A}} & @USER he fucking kills me. he knew it was coming & 0.919 & 0.958 & 0.852 & 0.509 & 0.809 & 0.177 & OFF & E \\
& His kissing days are over, he's a pelican now! & 0.659 & 0.304 & 0.568 & 0.523 & 0.514 & 0.131 & NOT & H \\
& i think we're all in love with winona ryder & 0.060 & 0.038 & 0.017 & 0.480 & 0.102 &	0.155 & NOT & E \\
\midrule
\multirow{3}{*}{\textbf{B}} & Guess I’ll just never understand the fucking dynamics & 0.901 & 0.569 & 0.001 & 0.617 & 0.522 & 0.327 & UNT & H \\ 
& @USER Government is a bunch of bitches. & 0.013 & 0.221 & 0.000 & 0.397 & 0.158 & 0.164 & TIN & E \\
& @USER Give me the date. Fuck them other niggas Bro I’m irritated as fuck & 0.882 & 0.666 & 0.983 & 0.701 & 0.808 & 0.131 & TIN & E \\
\midrule
\multirow{3}{*}{\textbf{C}} & @USER He was useless stupid guy & 0.807 & 0.915 & 1.000 & 0.480 & 0.801 & 0.197 & IND & E \\
& It's like mass shootings is the reg in this shit hole country! & 0.826 & 0.479 & 0.693 & 0.570 & 0.642 & 0.131 & OTH & H \\
& Getting these niggas tatted is a overstatement are ya dead serious & 0.700& 0.691 & 0.770 & 0.491 & 0.663 & 0.104 & GRP & H\\
\bottomrule
\end{tabular}
\caption{Training data aggregation examples. Columns 3-6 show the confidence of each of the models with respect to the positive class in Levels A and B (OFF, UNT) and only for the corresponding class in C (one example for each of the classes: TIN, GRP, OTH). The \emph{label} column shows manual annotations, and the last column shows whether the tweet is considered \emph{Easy} (\emph{E}) or \emph{Hard} (\emph{H}) based on its average (\emph{AVG}) confidence. \emph{FT} stands for FastText.}
\label{tab:training-data-examples}
\end{table*}

The decrease in performance in the final level can lead to increased noise in the semi-supervised labels, but we use an ensemble of four models, and we provide the average and the standard deviation of the confidence across the models on each instance to mitigate this. As we show later, these scores can be successfully used to filter out a large amount of noise in the semi-supervised dataset, thus yielding performance improvements.

We computed the aggregated single prediction based on the average and the standard deviation of the confidences predicted by each of the models: $\olidweak{} = \{(x_i',p_i')|i \in [1, |\olidweak|])\}$, where ${p'_i} = \mathrm{avg}(\{{p'_i}^j|j \in [1, N]\})$. In particular, we computed the scores based on the confidences for the positive class at Levels A and B, and on the confidences for IND, GRP, and OTH at Level C. We performed the above aggregation step instead of just using the scores of each model in order to avoid over-fitting to any particular model in the ensemble. This also helps to prevent biases with respect to individual models in future uses of the dataset. Moreover, the standard deviation and the average scores can be used to filter instances that the models disagree on, thus reducing the potential noise in the semi-supervised annotations.

We labeled the dataset in this semi-supervised manner by first assigning a Level A label to all the tweets. Then, we selected the subset of tweets that were likely to be offensive for all models (BERT and LSTM $\geq.5$, PMI and FT=OFF) as instances that should be assigned a label for Level B. Finally, for Level C, we chose the tweets that were likely to be TIN at Level B with a standard deviation lower than 0.25.

Thus, only the instances that were most likely to be offensive were considered at Levels B and C, and only those that were most likely to be offensive and targeted were considered at Level C. The size and the label distribution across the datasets can be found in Table~\ref{tab:training-data-distribution} and examples of tweets along with model prediction confidences can be found in Table~\ref{tab:training-data-examples}.

\begin{table*}[t]
\centering
\small
\begin{tabular}{ccclll}
\toprule
\bf \# & \textbf{Type} & \textbf{Prediction} & \textbf{Tweet} & \bf Gold Label\\
\midrule
1 & Easy & OFF & this job got me all the way fucked up real shit & OFF UNT \\
2 & Easy & OFF & @USER It’s such a pain in the ass & OFF UNT\\
3 & Easy & OFF & wtf ari her ass tooo big & OFF TIN IND\\
\midrule
4 & Easy & NOT & This account owner asks for people to think rationally. & NOT\\
\midrule
5 & Hard & OFF & It sucks feeling so alone in a world full of people & NOT\\
6 & Hard & OFF & @USER We are a country of morons & OFF TIN GRP\\
\midrule
7 & Hard & NOT & Hate the sin not the sinner... & NOT\\
8 & Hard & NOT & Somebody come get her she's dancing like a stripper & OFF TIN IND\\
\bottomrule
\end{tabular}
\caption{Example tweets from the \olidweak{} \emph{test} dataset and its four subsets. Shown are the difficulty of each subset (\emph{Type}), the ensemble model prediction for the examples in each subset (\emph{Prediction}), an example tweet's text, and the manually annotated gold label.}
\label{tab:analysis:examples}
\end{table*}

\subsection{\olidweak{} Test Dataset}

As the \olid{} test set was very small, particularly for Levels B and C, we also annotated a portion of our held-out three million tweets in order to create a new \olidweak{} test set to obtain more stable results and to analyze the performance of various models in more detail. 

First, all co-authors of this paper (five annotators) annotated 48 tweets that were predicted to be OFF in order to measure inter-annotator agreement (IAA) using $P_0=\frac{agreement\_per\_annotation} {total\_annotations*num\_annotators}$. We found IAA to be \textit{0.988} for Level A; an almost perfect agreement for OFF/NOT. The IAA for Level B was \textit{0.818}, indicating a good agreement on whether the offensive tweet was TIN/UNT. Finally, for Level C, the IAA was \textit{0.630}, which is lower but still considered reasonable, as Level C is more complicated due to its 3-way annotation schema: IND/GRP/OTH. Moreover, while a tweet may address targets of different types (e.g.,~both an individual and a group), only one label can be chosen for it.

After having observed this high IAA, we annotated additional offensive tweets with a single annotation per instance. We divided our Level A data into four portions based on model confidence: 

\begin{itemize}
    \item \textit{if} BERT $\!\geq\! .8$ $\land$ PMI $\!=\!$ OFF $\land$ FT $\!=\!$ OFF $\land$ LSTM $\!\geq\! .8$ \textit{then} \textbf{\textit{Easy} OFF} [2,380 tweets]
    \item \textit{else if} BERT $\geq$ .5 $\!\land\!$ PMI $\!=\!$ OFF $\land$ FT $\!=\!$ OFF $\land$ LSTM $\!\geq\! .5$ \textit{then} \textbf{\textit{Hard} OFF} [835 tweets]
    \item \textit{else if} BERT $\! \leq\! .2$ $\land$ PMI $\!=\! $ NOT $\land$ FT $\!=\!$ NOT $\land$ LSTM $\! \leq\! .8$ \textit{then} \textbf{\textit{Easy} NOT} [2,500 tweets]
    \item \textit{else if} BERT $\!<.5\!$ $\land$ PMI $\!=\!$ NOT $\land$ FT $\!=\!\!$ NOT $\land$ LSTM $\!<\! .5$ \textit{then} \textbf{\textit{Hard} NOT} [278 tweets]
\end{itemize}

Note that PMI $\!=\!$ OFF and FT $\!=\!$ OFF designates that the model's probability is higher for OFF than for NOT. We selected the rest of the thresholds after a manual examination of the confidence scores for each model. We chose the threshold where the model is confident and mostly correct.

We annotated 3,493 tweets for Level A. The number of annotations at each level is shown above in square brackets. Moreover, in order to create a complete test dataset for Level A (where we only labeled offensive tweets), we also took a random set of 2,500 \textit{Easy} NOT tweets. The resulting test sizes are shown in Table~\ref{tab:training-data-distribution}. Of the 3,493 annotated tweets, 491 were judged to be NOT. In total, there were 5,993 tweets in our test set. In all cases, we annotated all three levels, but the decision about whether a tweet in Level B/C is \textit{Easy} or \textit{Hard} is still based on its Level A confidence.

Table~\ref{tab:analysis:examples} shows some tweets and whether they are \textit{Easy} OFF/NOT (lines 1-4) or \textit{Hard} OFF/NOT (lines 5-8), and Table~\ref{tab:analysis} shows statistics about the \emph{Easy} and the \emph{Hard} examples in the test dataset. Note that determining the labels for the \textit{Hard} examples is not simple and the model does make incorrect predictions such as in lines 5 and 8 of Table~\ref{tab:analysis:examples}. In fact, 25\% of the \textit{Hard} OFF tweets that we annotated were NOT. In contrast, 8\% of the \textit{Easy} OFF tweets were judged to be NOT. 

\begin{table}[t]
\centering
\small
\begin{tabular}{ccrrr}
\toprule
\multirow{2}{*}{\textbf{Type}} & \bf Model & \multicolumn{2}{c}{\bf Gold Label} & \bf \multirow{2}{*}{Total}\\ & \textbf{Prediction} & \textbf{OFF} & \textbf{NOT}\\
\midrule
easy & OFF & 2,187 & 193 & 2,380\\
easy & NOT & 0 & 2,500 & 2,500\\
hard & OFF & 670 & 165 & 835\\
hard & NOT & 145 & 133 & 278\\
\midrule
\multicolumn{2}{l}{\bf Total} & \bf 3,002 & \bf 2,991 & \bf 5,993\\
\bottomrule
\end{tabular}
\caption{Statistics about the \olidweak{} \emph{test} dataset grouped by difficulty (\emph{Type}) and model prediction.}
\label{tab:analysis}
\end{table}

\section{Experiments and Evaluation}
\label{sec:results}

Below, we describe our experiments and evaluation results on the \olid{} test set when training on \olid{} + \olidweak{} compared to training on \olid{} only.

\subsection{Experimental Setup}
\label{subsec:experiment_setup}

We used the BERT and the FastText models from the semi-supervised annotation setup to estimate the improvements when training on the supervised dataset \olid{} together with the semi-supervised \olidweak{}. The models in all sets of experiments were fine-tuned on a 10\% validation split of the training set used during co-training. We explored different ways to combine \olid{} and \olidweak{}, and different thresholds for the confidence of the instances in \olidweak{}. We achieved improvements for Levels B and C by upsampling the underrepresented classes: we sampled $K$ instances of each class, where $K$ is the number of instances for the most frequent class. We also removed the warm-up in Levels B and C, which improved the results further.

\paragraph{FastText.} The FastText model is implemented as an external command-line tool, which does not give us much control over training. Thus, we trained models on the combined training sets of \olid{} and \olidweak{}. The FastText model had the same parameters used above in co-training.

\begin{table}[t]
\centering
\small
\begin{tabular}{c@{ }c@{ }c@{ }c@{ }c@{ }c@{ }}
\toprule
\multirow{2}{*}{\textbf{Level}} & \multirow{2}{*}{\textbf{Baseline}} & \multicolumn{2}{c}{\bf BERT} & \multicolumn{2}{c}{\bf FastText}\\
&& \textbf{\olid} & \textbf{+\olidweak{}} & \textbf{\olid} & \textbf{+\olidweak{}}\\ \midrule
A & 0.419 & \textbf{0.816} & 0.809 & 0.662 & \textbf{0.720}\\
B & 0.470 & 0.687 & \textbf{0.729} & 0.470 & \textbf{0.591} \\ 
C & 0.214 & 0.589 & \textbf{0.643} & \textbf{0.590} & 0.515\\ 
\bottomrule
\end{tabular}
\caption{Evaluation results on the \olid{} \emph{test} dataset (macro-F1 scores) for BERT and FastText with and without training on \olidweak{}, compared to the majority class baseline.}
\label{tab:results}
\end{table}

\paragraph{BERT.} Due to the computational requirements of BERT, we subsampled 20,000 tweets from \olidweak{} in Levels A and B; in fact, using more instances did not help. During training, we used \olidweak{} in the first epoch and \olid{} in the following two epochs for Level A. Using \olidweak{} after training with \olid{} yielded worse results, which is probably because the semi-supervised dataset by construction contains somewhat noisy labels. Yet, it can be used as an initial step to guide the model towards a better local minimum. On the other hand, we conjecture that the supervised dataset is better suited for fine-tuning the model towards the local minimum with the gold data, particularly in Level A, where the training split of \olid{} is already sufficient for training BERT. For Levels B and C, we trained for two epochs with the training split of \olid{} and then for one epoch with \olidweak{}. At Levels B and C, we observed that training with \olidweak{} in the first epochs and then fine-tuning with \olid{} did not improve the performance. Moreover, training with \olid{} and then using \olidweak{} for the final epochs yielded substantial performance improvements. We assume this is due to the small training size of \olid{}, which can cause the model to overfit to a suboptimal local minimum when used in the final training epochs.

\paragraph{Selecting \olidweak{} Instances.}  We filtered the training instances from \olidweak{} to be the most confident examples based on the average probability score provided in \olidweak{} when training using FastText and BERT.
We chose the threshold for the average confidence score based on the validation dataset as follows:

{\bf Level A:} $avg($OFF$)\!<\!0.20\, \cup\, avg($OFF$)\!>\!0.70$

{\bf Level B:} $avg($UNT$)\!<\!0.35\, \cup\, avg($UNT$)\!>\!0.65$

{\bf Level C:} $avg($IND$)\!>\!0.80\, \cup\, avg($GRP$) \!>\!0.70\, \cup\, avg($OTH$)\!>\!0.65$

We selected the labels as follows: in Level A, NOT when $avg($OFF$)<0.20$, else OFF; in Level B, UNT when $avg($UNT$) > 0.65$, else TIN; in Level C, the class with the highest probability.

\subsection{\olid{} Results}

In this section, we describe our results when testing on the \olid{} test set. We compare training on \olid{} vs. training on \olid{} + \olidweak{}. The results are shown in Table~\ref{tab:results}.

We can see that for Level A, when training with \olid{}+\olidweak{}, the results improve for FastText, which is a weak model (see also Table~\ref{tab:ensemble}). However, for BERT, which already performs very strongly when fine-tuned with \olid{} only, there is not much difference when \olidweak{} is added; in fact, there is even a small degradation in performance. These results are in line with findings in previous work \cite{longstaff2010improving}, where it was observed that democratic co-training performs better when the initial classifier accuracy is low.

For Level B, the \olid{} training dataset is smaller, and the task is more complex. Thus, there is more benefit in adding \olidweak{}, which yields sizable improvements for both BERT and FastText. Yet, as FastText is a much weaker model (in fact, performing the same as the majority class baseline when trained on \olid{} only), the absolute gain for it is much larger than for BERT: 12.1 vs. 4.2 macro-F1 points absolute.

Finally, for Level C, the manually annotated \olid{} dataset is even smaller, and the number of classes increases from two to three. As a result, BERT benefits from adding the \olidweak{} data by a large margin of 5.4 macro-F1 points absolute. However, using \olidweak{} for FastText does not help. This might be due to FastText already achieving high performance when trained with \olid{} only (see Table~\ref{tab:ensemble}), which is on par with that of BERT, while democratic co-training performs well when the initial classifier's performance is low.

\subsection{\olidweak{} Results}
\label{sec:discussion}

Above, we have demonstrated sizable improvements when training on a combination of the \olid{} and the \olidweak{} datasets, and testing on the test part of \olid{}. However, \olid{} is small, and thus the results could be unstable, especially for Levels B and C. Thus, evaluating on a larger set, namely the test set of \olidweak{}, is important for estimating the model stability. We also focus on \emph{Easy} vs. \emph{Hard} examples (based on the confidence computed during co-training) to gain better insight into why some tweets are easier to classify as offensive than others. The results are shown in Table~\ref{tab:results-new} and they beat the majority class baselines by a huge margin.

\begin{table}[t!]
\centering
\small
\begin{tabular}{@{ }c@{ }@{ }@{ }lc@{ }c@{ }@{ }cc@{ }@{ }c}
\toprule
 & \multirow{2}{*}{\textbf{Model}} & \multirow{2}{*}{\textbf{Baseline}} & \multicolumn{2}{c}{\bf BERT} & \multicolumn{2}{c}{\bf FastText} \\
 & & & \olid{}& +\olidweak{} & \olid{}& +\olidweak{} \\
\midrule
\multirow{3}{*}{A} & Full & 0.338 & 0.922 & \bf 0.923 & 0.856 & \bf 0.860 \\
 & Easy & 0.400 & 0.983 &\bf 0.983 & 0.936 & \bf 0.940 \\
 & Hard & 0.444 & 0.557 & \bf 0.570 & 0.525 & \bf 0.536 \\
\midrule

\multirow{3}{*}{B} & Full & 0.236 & 0.559 &\bf 0.666 & 0.355 & \bf 0.493\\  & Easy & 0.232 & 0.569 &\bf 0.677 & 0.349 & \bf 0.509\\
 & Hard & 0.234 & 0.542 & \bf 0.649 & 0.363 & \bf 0.467 \\
\midrule
\multirow{3}{*}{C} & Full & 0.203 & 0.627 & \bf 0.645 & 0.387 &\bf 0.504\\  & Easy & 0.201 & 0.635 & \bf 0.644 & 0.378 &\bf 0.504 \\
 & Hard & 0.205 & 0.616 &\bf 0.649 & 0.397 &\bf 0.505 \\
\bottomrule
\end{tabular}
\caption{Evaluation results on the \olidweak{} \emph{test} dataset (macro-F1 scores), and on its \emph{Easy} and \emph{Hard} subsets, compared to the majority class baseline.}
\label{tab:results-new}
\end{table}

We can see that the results for Level A on \olidweak{} test are 0.923 and 0.860 macro-F1 for BERT and for FastText, respectively, with a small improvement when \olid{} is augmented with \olidweak{} for FastText only. This is consistent with what we found on the \olid{} test set. Note that the full results for Level A are much better than those on the OLID test dataset in Table~\ref{tab:results}. We believe that this is partially due to our selection of tweets for the new test set, indicating that there are more \textit{Easy} tweets in it. Similar findings to the full test set occur with the \emph{Easy} tweets, but the scores this time are even higher. On the other hand, for the \emph{Hard} tweets, the results are much lower at 0.570 and 0.536 for BERT and for FastText, respectively. Overall, using \olidweak{} yields a nice improvement for both models on the \emph{Hard} tweets, which was not evident in the \olid{} test set in Table~\ref{tab:results}. 

In order to gain further insight into why the results are so high for \emph{Easy} OFF tweets at Level A, we implemented a curse-word baseline using the absence and the presence of 22 curse words (the list can be found in Appendix~\ref{App:A}). 
We found that most \textit{Easy} tweets were classified correctly by this baseline with an F1-score of 0.936. In contrast, the curse-word baseline was not effective on the hard examples, just like the BERT and the FastText models were not. It achieved a macro-F1 score of 0.580, which is one point higher than the BERT result. Thus, we can conclude that both BERT and FastText are probably overfitting to the curse words to some extent. The \textit{Hard} tweets are offensive due to other language use such as negative biases rather than the appearance of a curse word such as in examples 6 and 8 in Table~\ref{tab:analysis:examples}. Classifying such tweets correctly remains an open challenge not only for our models, but also in general.

The difference between \emph{Easy} OFF/NOT and \emph{Hard} OFF/NOT tweets is less pronounced for Levels B and C. The curse word imbalance may have a small impact on the lower levels as UNT tweets are more likely to contain curse words. In all cases, combining \olidweak{} and \olid{} for Levels B and C yields a sizable improvement, indicating that the larger test set can better showcase the differences, leading to more stability. The results for Levels B and C vary greatly for the two models compared to those on the \olid{} test set in Table~\ref{tab:results}, which points to the challenges of having a small test set.

\section{Conclusion and Future Work}
\label{sec:conclusion}

We have presented \olidweak{}, a large-scale semi-supervised training dataset for offensive language identification, which we created using an ensemble of four different models. To the best of our knowledge, \olidweak{} is the largest dataset of its kind, containing nine million English tweets. We have shown that using \olidweak{} yields noticeable performance improvements for Levels B and C of the OLID annotation schema, as evaluated on the \olid{} test set. Moreover, in contrast to using keywords, our approach allows us to distinguish between \textit{Hard} and \textit{Easy} offensive tweets. The latter enables us to have a deeper understanding of offensive language identification and indicates that detecting \textit{Hard} offensive tweets is still an open challenge. Our work encourages safe and positive places on the web that are free of offensive content, especially non-obvious cases, i.e.,~\textit{Hard}. \olidweak{} was the official dataset of the SemEval shared task OffensEval 2020 \cite{zampieri-etal-2020-semeval}. 

In the future, we would like to provide insights and methods for categorizing \textit{Hard} tweets.

\section*{Acknowledgements} 
We would like to thank the anonymous reviewers for their constructive comments. This work is part of the Tanbih mega-project, developed at the Qatar Computing Research Institute, HBKU, which aims to limit the impact of ``fake news,'' propaganda, and media bias by making users aware of what they are reading.
Pepa Atanasova has received funding from the European Union's Horizon 2020 research and innovation programme under the Marie Skłodowska-Curie grant agreement No 801199.
\begin{figure}[h!]
\includegraphics[width=40pt]{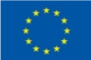}
\end{figure}

\section*{Ethics Statement}

\paragraph{Dataset Collection} We collected both the \olid{} and the \olidweak{} datasets using the Twitter API. The \olid{} dataset was collected using keywords that would be more likely to be accompanied by offensive tweets \cite{zampieri-etal-2019-predicting}, while the \olidweak{} dataset was collected by querying with frequent stop words (see Section \ref{S:dataset}).
Overall, we followed the terms of use outlined by Twitter.\footnote{\url{http://developer.twitter.com/en/developer-terms/agreement-and-policy}} Specifically, we only downloaded public tweets, and we provided only the user ids of those tweets in order to ensure that deleted tweets will no longer be part of our dataset. Moreover, in all our examples in this paper, we anonymized the user names in the tweets. Since no private information is stored, IRB approval is not required. All annotations were performed internally by the authors of the paper.

\paragraph{Biases} We note that determining whether a piece of text is offensive can be subjective, and thus it is inevitable that there would be biases in our gold labeled data. It is expected that such biases will, therefore, also be present in the semi-supervised dataset we generated from such tweets.

While we cannot ensure that no biases occur in the gold data, we addressed these concerns by following a well-defined schema, which sets explicit definitions for offensive content during annotation. Our high inter-annotator agreement makes us confident that the assignment of the schema to the data is correct most of the time.

Using semi-supervised techniques to create a large dataset, \olidweak{}, can cause the biases found in the gold data to be expanded further. We mitigated this in two ways. First, we gathered tweets based on the most frequent words in English to ensure a random set of initial tweets. Next, we constructed an ensemble of models with diverse inductive biases to label the target tweet, which can help to ameliorate the individual model biases and to produce predictions with a lower degree of noise. At test time, we aimed to have a meaningful ratio of offensive and non-offensive tweets based on a random collection of tweets. We also labeled all testing offensive tweets manually. The aim of these steps was to help reduce the potential biases. Please refer to Section~\ref{App:C} of the Appendix for some analysis that shows the diversity of the models.

We acknowledge that current semi-supervised techniques do not address the problem of potential biases, which is inherent in the semi-supervised data coming from the supervised source model(s), which can also be studied in future work. We further acknowledge that biases can still exist in the ratio of offensive to non-offensive tweets in our dataset. In general, the size and the method of collection for the \olidweak{} dataset mean that biases are hard to avoid.

Moreover, offensive language can vary depending on demographics, such as the gender of the targeted individual; the target could even be a particular gender group. Such biases, which are present in natural language data~\cite{Olteanu2019SocialDB}, are an important direction for future work. 

\paragraph{Misuse Potential} Most datasets compiled from social media present some risk of misuse. We therefore ask researchers to be aware that the \olidweak{} dataset can be maliciously used to unfairly moderate text (e.g.,~a tweet) that may not be offensive based on biases that may or may not be related to demographic and/or other information present within the text. Intervention by human moderators would be required in order to ensure that this does not occur. 

\paragraph{Intended Use} We have presented the \olidweak{} dataset with the aim to encourage research in automatically detecting and stopping offensive content from being disseminated on the web. Such systems can be used to alleviate the burden for media moderators, which can suffer from psychological disorders due to the exposure of extremely offensive content. Improving the performance of offensive content detection systems can decrease the amount of work for human moderators, but human supervision would still be required for more intricate cases and in order to ensure that the system is not causing harm. With the possible ramifications of a highly subjective dataset, we distribute \olidweak{} for research purposes only, without a license for commercial use. Any biases found in the dataset are unintentional, and we do not intend to cause harm to any group or individual.

We believe that this dataset is a useful resource when used in the appropriate manner and that it has great potential to improve the performance of current offensive content detection and automatic content moderation systems. 

\bibliography{custom}
\bibliographystyle{acl_natbib}

\newpage

\appendix
\section{Appendix}

Below, we provide additional details about the data collection, we perform analysis, and we give some implementation details.

\subsection{Data Collection and Analysis}
\label{App:A}

In Section~\ref{sec:data:tweets}, we described our method for collecting tweets. We queried the Twitter API using the most frequent English words based on the large monolingual Project Gutenberg corpus.\footnote{\url{ http://en.wiktionary.org/wiki/Wiktionary:Frequency_lists\#Project_Gutenberg}} Table~\ref{tab:freq_words} shows the top-20 most frequent words in the corpus and their frequency, which we used to collect the tweets. The normalized value is the percentage of the total frequency for the first $N$ most frequent words. To choose a word, we generate a random number between 0 and 1, and we select the word corresponding to the smallest number that is higher than the generated one.
For example, 0.45 would correspond to the word \emph{to}.

\begin{table}[tbh]
\centering
\scalebox{.83}{
\begin{tabular}{lrc|lrc}
\toprule
\bf w & \bf Frequency & \bf Norm. & \bf w & \bf Frequency & \bf Norm. \\
\midrule
the 	&	56,271,872	&	0.20	& it 	&	8,058,110	&	0.79	\\
of 	&	33,950,064	&	0.32 & with 	&	7,725,512	&	0.82		\\
and 	&	29,944,184	&	0.43 & is 	&	7,557,477	&	0.85	\\
to 	&	25,956,096	&	0.52 & for 	&	7,097,981	&	0.87	\\
in 	&	17,420,636	&	0.58 & as 	&	7,037,543	&	0.90	\\
i 	&	11,764,797	&	0.63 & had 	&	6,139,336	&	0.92	\\
that 	&	11,073,318	&	0.67 & you 	&	6,048,903	&	0.94	\\
was 	&	10,078,245	&	0.70 & not 	&	5,741,803	&	0.96	\\
his 	&	8,799,755	&	0.73 & be 	&	5,662,527	&	0.98	\\
he 	&	8,397,205	&	0.76 & her 	&	5,202,501	&	1.00	\\
\bottomrule
\end{tabular}
}
\caption{The top-20 most frequent English words ({\em w}). {\em Norm.} is the normalized value based on the total frequency of the top words.}
\label{tab:freq_words}
\end{table}

In Section~\ref{sec:discussion}, we discussed the simple curse-word baseline used to analyze the \textit{Easy} OFF/NOT tweets. Table~\ref{tab:curse_words} gives the list of the 22 curse words that we used in that baseline.

\begin{table}[tbh]
\centering
\scalebox{.85}{
\begin{tabular}{lllll}
\toprule
 ass & arse & wtf & lmao & fuck \\
  bitch & nigga & nigger & cunt & effing \\
  shit &  hell & damn &  crap & bastard \\
  idiot & stupid & racist & dumb & f*ck  \\
 pussy & dick & & \\
 \bottomrule
 \end{tabular}
 }
 \caption{The 22 common offensive terms used in the curse-word baseline.}
 \label{tab:curse_words}
 \end{table}

\subsection{Implementation Details}
\label{App:C} 

We fine-tuned the models on 10\% of the \olid{} dataset. All models were trained on an NVIDIA Titan X GPU with 8GB of RAM. The performance of the individual models in our ensemble for semi-supervised labelling is shown in Table~\ref{tab:valensemble}. The evaluation measure we used for all experiments is macro-F1 score, as implemented in scikit-learn.\footnote{\url{scikit-learn.org/stable/modules/generated/sklearn.metrics.f1_score.html}}

\begin{table}[tbh]
\centering
\scalebox{.85}{
\begin{tabular}{lrrr}
\toprule
\textbf{Model} & \textbf{A} & \textbf{B} & \textbf{C} \\\midrule
BERT & 0.788 & 0.610 & 0.577 \\
PMI & 0.772 & 0.595 & 0.536  \\
LSTM & 0.599 & 0.599 & 0.579 \\
FastText & 0.672 & 0.489 & 0.456 \\
\bottomrule
\end{tabular}
}
\caption{Macro-F1 score, on the validation set, for the models used in the ensemble for Levels A, B, and C.}
\label{tab:valensemble}
\end{table}

In Table~\ref{tab:model_agreement}, we show the agreement between the models for the task prediction. For Levels A and B, it is more common that all four models agree, while in Level C, there are more cases when at least one model disagrees with the rest. Moreover, in Level A, there are almost no cases when the decision is tied with two models disagreeing with the other two. Finally, as in Level C the performance of the models is lower, the disagreement between the models in the ensemble is the largest and it is least common for all four models to agree on a prediction. Given the observed agreement rates, we conclude that there is considerable variance in the predictions across the models, especially for the lower levels. This indicates that the individual models can have differences in their predictions, which can be resolved by the ensemble combination in the democratic training setup.

\begin{table}[tbh]
\centering
\scalebox{.85}{
\begin{tabular}{lccc}
\toprule
\textbf{N} & \textbf{A} & \textbf{B} & \textbf{C} \\\midrule
4 & 0.517 & 0.598 & 0.249 \\
3 & 0.392 & 0.275 & 0.417  \\
2 & 0.091 & 0.127 & 0.335 \\
\bottomrule
\end{tabular}
}
\caption{Percentage of instances where $N$ models agree on a predicted label of an instance, $N \in \{2, 3, 4\}$, for Levels A, B, and C. }
\label{tab:model_agreement}
\end{table}

\end{document}